\documentclass[letterpaper]{article} 
\usepackage{aaai23}  
\usepackage{times}  
\usepackage{helvet}  
\usepackage{courier}  
\usepackage[hyphens]{url}  
\usepackage{graphicx} 
\usepackage{natbib}  
\usepackage{caption} 
\usepackage{algorithm}
\usepackage{algorithmic}

\usepackage{newfloat}
\usepackage{listings}
\DeclareCaptionStyle{ruled}{labelfont=normalfont,labelsep=colon,strut=off} 
\floatstyle{ruled}
\newfloat{listing}{tb}{lst}{}
\floatname{listing}{Listing}
\nocopyright 

\title{Learning and Leveraging Verifiers to Improve Planning Capabilities of Pre-trained Language Models}
\author {
    Daman Arora,\textsuperscript{\rm 1}
    Subbarao Kambhampati \textsuperscript{\rm 2}
}
\affiliations {
    \textsuperscript{\rm 1} Indian Institute of Technology, Delhi\\
    \textsuperscript{\rm 2} Arizona State University\\
    cs5180404@iitd.ac.in, rao@asu.edu
}

\usepackage{bibentry}

\begin{document}

\maketitle

\begin{abstract}
There have been wide spread claims in the literature about the emergent reasoning capabilities of Pretrained Large Language Models. However, recent studies, have found that their ability to plan remains questionable. Through our experiments using GPT-2, we empirically demonstrate that the performance of a finetuned baseline remains poor because it violates pre-conditions of actions in the plans that it generates. To improve the planning capabilities of a finetuned LLM, we train a verifier, which can classify actions as being valid or invalid in a particular state. By randomly sampling actions from the same dataset, we generate examples of invalid actions which are then used to train a verifier which can check for action applicability. In the presence of diverse sampling from a generator and a verifier which can prune invalid trajectories, we show significant gains in the success rate on the Blocksworld domain. Additionally, we show that finetuning the GPT-2 generator itself to create the verifier generalizes better than finetuning the base GPT-2. Lastly, we investigate the role of the sampling temperature which can be used to control the exploration-exploitation tradeoff. 
\end{abstract}

\section{Introduction}

Pre-trained Large Language Models which employ Transformers\cite{vaswani2017attention} as the backbone architecture have made huge strides in the field of Deep Learning and become the de-facto architecture for a variety of tasks in NLP. These models posses a rich amount of knowledge about the world, as a by-product of their pre-training. Recent works have hinted at their emergent abilities to reason \citep{kojima2023large}. Sometimes, the reasoning can be elicited by prompting the model in the correct fashion, referred to as Chain-of-Thought Prompting\cite{wei2023chainofthought}. 

This leads to a natural question: can these models plan? A recent work \cite{llmPlanRao} investigates the planning abilities of various LLMs like GPT-3, InstructGPT and BLOOM. In this work, the authors prompt the model with a textual description of the domain and an example plan, in hope that the LLM could generalize to other instances. However, the results show very poor performance on test instances. Even after finetuning the models on plans generated by a SOTA planner, the performance remains sub-par.

The PDDL framework provides a convenient abstraction for planning by specifying relations over objects of different types. Actions are parameterized by objects and have pre-conditions and effects. 
The ability of an LLM like architecture to plan subsumes the ability to build a world-model of a domain. In the context of PDDL, knowing a world-model would be equivalent to knowing the effects and pre-conditions of an action. Our investigation of the poor performance of finetuned models reveals that they don't perform well because eve the finetuned models often take actions which are invalid in a particular state, leading to failed plans. This leads to poor success rate of such planners on test instances. In other words, when trained on trajectories, LLMs are bad at capturing pre-conditions of actions. 

In order to alleviate this problem of finetuned models, we train a verifier which classifies a generated action as valid or invalid in a given state. Verifiers have been proposed in the context of reasoning and problem-solving for Language Models. For instance, \cite{verifier_math} train a verifier for mathematical problem solving in natural language by explicitly labeling solutions generated by a finetuned generative language model. The ``good'' examples are present in the finetuning dataset itself. However, the ``bad'' examples requires human annotation which can be cumbersome and expensive. Along similar lines, Yang et al propose verifier-guided search\cite{yang2022generating}, where each proof step is a (premise, conclusion) pair. For training a verifier, they propose augmentations, such as negating the conclusion, removing parts of the premise, etc.

In the context of PDDL planning, we propose a cheap source of negative examples. For a state in the dataset, we generate a negative example by sampling any random action present in the dataset. Equipped with this simple strategy, we can train a verifier. We demonstrate that sampling a diverse set of trajectories and using the verifier to prune them is critical for improving the success rate of these planners. 

For our experiments, we use Blocksworld as the testbed and finetuned GPT-2\cite{gpt2} models as the generator and the verifier. Results show that given the same number of attempts a generator equipped with a verifier achieves a gain of $27\%$ over the pure-generator baseline. Additionally, the rate of bad trajectories decreases from $52\%$ to a mere $5\%$. We also investigate the role of the sampling temperature of the generator and demonstrate it's role in balancing the exploration-exploitation tradeoff. Finally, we show that finetuning the generator itself for verification generalizes better as compared to finetuning vanilla GPT-2. 


\section{Verifier Augmented Plan Generation in LLMs} 

The PDDL framework \cite{pddl} defines a class of planning problems where each problem can be defined using a domain $\mathcal{D}$, a start state $\mathcal{I}$ and a goal specification $\mathcal{G}$. The domain consists of a set $\mathcal{F}$ of predicates and a set $\mathcal{A}$ of action templates. An instantiation of the domain consists of grounding the predicates over a set of objects $\mathcal{O}$.  A grounded action consists of an action template $a$  and a mapping $\mathcal{V}$ of each argument of an action to an object. Such a mapping induces a set of pre-conditions $pre(a[\mathcal{V}])$ and a set of effects $eff(a[\mathcal{V}])$. A grounded action is said to be applicable in a state if all its preconditions are true. When a grounded action is applied to a state, all the effects of the action become true in the next state. 

We aim to take a \textbf{generate-test }loop approach to planning  using LLMs, where the \textbf{generate} part is a finetuned LLM and the \textbf{test} part is also a finetuned LLM. We assume an offline dataset of successful trajectories $\{(s_1, a_1, s_2, a_2..., s_g)\}$ where each $s_i$ contains a textual representation of the state.  The generator and verifier are both learnt using the same dataset. Since the domain is Markovian and we work with goal-conditioned policies, policy learning can be formulated as an auto-regressive generation task wherein the model has to generate an action and the corresponding next state given a state and the goal. For this purpose, we finetune a pre-trained LLM GPT-2 \citep{gpt2} to be the generator. The generator is an LLM which is finetuned to generate the action and the next state $a_i, s_{i+1}$ given the goal state and the previous state $s_g, s_i$. The verifier is also a LLM which is finetuned to predict whether the action $a_i$ is applicable in state $s_i$.



\begin{figure}[ht]
\includegraphics[width=3.5in]{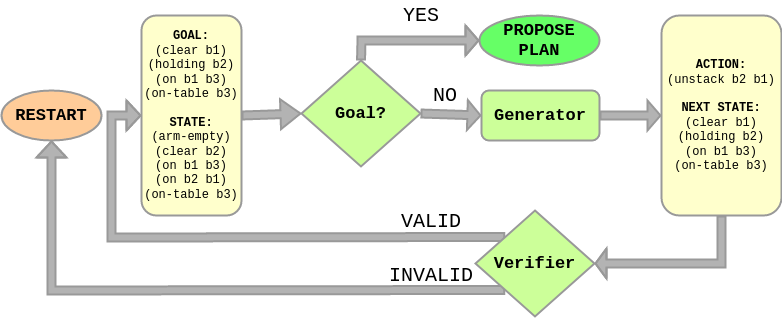}
    \caption{The figure shows a prompt which contains $s_i$ and $s_g$, which is given to the generator as input and the output it generates, which contains $a_i$ and $s_{i+1}$. The verifier taken the current state $s_i$ and generated action $a_i$ as input. After each transition generation, if the verifier approves the action, the generated next state $s_{i+1}$, is given as input for the next transition along with the goal $s_g$. If the verifier disapproves of the action, planner restarts from the initial state $s_0$.}
    \label{fig:prompt}
\end{figure}

\subsection{Training the Generator}
To create the generator, we finetune the pretrained GPT-2 model on transitions extracted from successful execution traces. From a trajectory $(s_1, a_1, s_2....s_g)$, where each $s_i$ and $a_i$ is the textual representation of the state and action respectively and $s_g$ is the goal, we train GPT-2 on a transition, containing $(s_g, s_i, a_i, s_{i+1})$. During inference, we provide with $(s_g, s_i)$ as the prompt and the model generates $(a_i, s_{i+1})$. \footnote{There are two reasons why it isn't preferred to include the entire history into the prompt. Firstly, the context window is of limited size. Secondly, if generation is done step by step, the inference time scales as $\mathcal{O}(n)$, whereas if the entire history is included in the prompt, it scales as $\mathcal{O}(n^2)$, $n$ being the \# of transitions.} An example of the prompt can be seen in Figure \ref{fig:prompt}.

\subsection{Training the Verifier}

In the context of planning, a verifier needs to learn which actions are applicable in a particular state. We hypothesize that preconditions can be implicitly learnt by a high-capacity model like GPT-2, given negative samples. Learning a verifier from successful trajectories is an interesting problem, since we do not have explicit knowledge about actions which are inapplicable in a particular state. We generate the training dataset for the verifier in the following manner: for every transition $(s_i, a_i, s_{i+1})$, $(s_i, a_i)$ is a positive sample. To generate a negative sample, we use $(s_i, a')$ where $a'$ is a random action samples from the dataset of trajectories. We note that $a'$ may well be executable in the state $s_i$, however, the probability of such an event is low. 

To train the verifier, we finetune the generator on the verification dataset. Since the GPT-2 architecture is a auto-regressive decoder architecture, the last token's contextual embedding contains the summary of the entire sentence, which is then passed through an MLP to perform a prediction task. In our case, the verifier is passed a state and an action, and the verifier has to do a classify the action as valid or invalid in the state. 

\subsection{Verifier-augmented Generation}

Figure~\ref{fig:prompt} shows the overall verifier augmented plan generation process. The GPT-2 Generator takes the initial state $s_0 \leftarrow s_I$ and goal $s_g$. At the $i^{th}$ transition the generator incrementally generates the next action $a_i$ and state $s_{i+1}$ from $s_i, s_g$. The verifier then checks whether the transition $s_i {\stackrel{a_i}{\rightarrow}} s_{i+1}$ is valid. If it is valid, then the generator continues with $s_{i+1}, s_g$.  If the transition is not valid, the generator is reset to start from $s_0, s_g$.\footnote{An alternative to resetting the generator all the way back to $s_0, s_g$ would be to just reset it to start from the previous state $s_i$. One difference is that $s_i, s_g$ may be an unsolvable planning problem even when $s_0, s_g$ is solvable.}  It should be noted that the learned verifier is not guaranteed to be correct in checking transition validity--it can have both false positives and false negatives.

\section{Experimental Evaluation}
\subsection{Generating the training data}
Using a randomized Blocksworld instance generator, with the same parameters, we generate 10,000 initial states with varying number of blocks and 200 test instances. To generate training trajectories, we use a \textbf{random-exploring}\footnote{Any planner can be used to generate such trajectories and is not important to demonstrate the utility of verifiers} agent which selects any action uniformly randomly from the set of applicable actions in a state. In order to generate some diversity, we prune actions in which the next state has already been explored. This gives us access to an offline dataset of valid and diverse trajectories $D = \{T_i=(s_1, a_1, s_2, a_2...s_g)\}_{i=1}^{|D|}$. To generate training data for GPT-2 from these trajectories, we choose the end state as the goal and each individual transition is taken as an input sentence\footnote{We only consider complete specification of the goal state in our work for simplicity, although a partial specification is an easy extension.}. Additional details can be found in the Appendix.

\subsection{Baselines and Metrics}
For all the methods, we consider a maximum plan length $L_{max}$. To measure the utility of a verifier we consider the following methods:
\begin{itemize}
    \item \textbf{generator@k}: This agent generates at most $k$ plans of length at most $L_{max}$. In the first plan that the agent's generated next state matches the goal, the plan is proposed. The inference algorithm is given in Algorithm \ref{alg:g@k}
    \item \textbf{generator+verifier@k}: This agent generates at most $k$ plans of length atmost $L_{max}$. As soon as a plan is found which reaches the goal and the verifier approves all transitions, it is proposed. The inference algorithm is given in Algorithm \ref{alg:g+v@k}
\end{itemize}

For all baselines, we use top-p sampling for the next token during generation \citep{toppsampling} with $p=0.99$, with a temperature $\tau=1$. $L_{max}$ is set to 40 steps.

Each planner proposes a plan for an instance. To assess the planners, we consider the following metrics:
\begin{itemize}
    \item \textbf{bad-transition-rate(BTR)}: Does the plan have an illegal action in the proposed plan?
    \item \textbf{goal-reaching-rate(GRR)}: Is the goal achieved in the proposed plan?
\end{itemize}

\begin{algorithm}[ht]
\caption{Inference algorithm for generator@k}
\begin{algorithmic}
\REQUIRE $G$: generator,$\;V$ : verifier, $L_{max}$: Max plan length, $k$: Max number of plans, $s_0$: Start state, $s_g$: Goal.
\STATE{$i \gets 0$}
\WHILE{$i < L_{max}$}
\STATE{$j\gets 0$}
\WHILE{$j < k$}
\IF{$s_{j} == g$}
\RETURN{$(a_1, a_2...a_{j-1})$}
\ENDIF
\STATE{$a_j, s_{j+1} \gets G(s_j, s_g)$}
\STATE{$j \gets j + 1$}
\ENDWHILE
\STATE{$i \gets i+1$}
\ENDWHILE

\end{algorithmic}
\label{alg:g@k}
\end{algorithm}

\begin{algorithm}[ht]
\caption{Inference algorithm for generator+verifier@k}
\begin{algorithmic}
\REQUIRE $G$: generator,$\;V$ : verifier, $L_{max}$: Max plan length, $k$: Max number of plans, $s_0$: Start state, $s_g$: Goal.
\STATE{$i \gets 0$}
\WHILE{$i < L_{max}$}
\STATE{$j\gets 0$}
\WHILE{$j < k$}
\IF{$s_{j} == g$}
\RETURN{$(a_1, a_2...a_{j-1})$}
\ENDIF
\STATE{$a_j, s_{j+1} \gets G(s_j, s_g)$}
\IF{$V(s_j, a_j)$ fails}
\STATE{\textbf{break}}
\ENDIF
\STATE{$j \gets j + 1$}
\ENDWHILE
\STATE{$i \gets i+1$}
\ENDWHILE

\end{algorithmic}
\label{alg:g+v@k}
\end{algorithm}

\subsection{Results}

\subsubsection{How does generator+verifier do?}

For the purpose of this experiment, we set $k=25$ and the trained model is used for generating plans on the 200 test instances. For each test instance, there can be three possible outcomes:

\begin{enumerate}
    \item The proposed plan is valid and reaches the goal.\footnote{An example plan can be found in the Appendix}
    \item The proposed plan fails because of illegal actions in any intermediate step. 
    \item The planner fails to propose a plan.
\end{enumerate}
\begin{table}[ht]
\centering
\begin{tabular}{ccc}
\textbf{Method}                & \textbf{GRR } & \textbf{BTR}    \\ \hline
generator@25     & 0.375 & 0.525 \\ \hline
generator+verifier@25 & 0.655 & 0.05
\end{tabular}
\caption{Results of generator@25 and generator+verifier@25 on 200 Blocksworld test instances.}
\label{tab:Results}
\end{table}
As we see from Table \ref{tab:Results}, generator@25 can only solve 37.5\% of the instances, and has a bad transition in 52.5\% of the total instances(Note that if even one action is inapplicable, the entire trajectory is invalid). However, generator+verifier@25 is able to gain substantially over generator@25 and achieve a score of 65.5\%. Additionally, we see that BTR goes down significantly and is present in only 5\% of the proposed plans in the test instances. 

\subsubsection{Does the performance improve with more attempts?}

\begin{figure}[!h]
    \centering
    \includegraphics[width=0.4\textwidth]{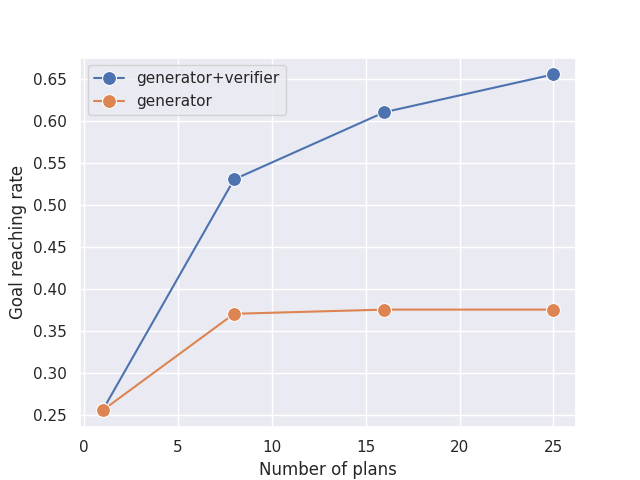}
    \caption{GRR of generator+verifier@k and generator@k with increasing number of attempts at generating a plan}
    \label{fig:GRR_num_plans}
\end{figure}

Since the generator is non-deterministic, we expect that the GRR of the planner would increase with increasing number of attempts that it gets at generating a plan. In order to verify this hypothesis, we vary $k$ in the range $\{1, 8, 16, 25\}$, and measure the performance of generator@k and generator+verifier@k. The results can be seen in Figure \ref{fig:GRR_num_plans}. It is evident that the performance of both models increases with increasing number of trajectories, but only generator+verifier@k is able to improve consistently unlike generator@k which plateaus around 16 trajectories.  

\subsubsection{Finetuning the generator to train the verifier leads to better generalization.}

Which model should be finetuned to create the verifier? The first option is to finetune GPT-2 itself, referred to as $V_{base}$. The other option is to finetune the generator, referred to as the $V_{generator}$\footnote{Note that the generator itself was finetuned from GPT-2.}. From results in  Table \ref{tab:Verifier_ablation}, we observe that finetuning the generator leads to better results. Intuitively, this could be because the generator has already been trained on a related task, which makes it more robust. 

\begin{table}[ht]
\centering
\begin{tabular}{ccc}
\textbf{Method}                & \textbf{GRR } & \textbf{BTR}    \\ \hline
generator+verifier($V_{base})$@25     & 0.635  & 0.105 \\ \hline
generator+verifier($V_{generator}$)@25 & 0.655 & 0.5 
\end{tabular}
\caption{Results of using a verifier finetuned on GPT-2 versus the generator.}
\label{tab:Verifier_ablation}
\end{table}

\subsubsection{How does the sampling temperature affect results?}
\begin{figure}[ht]
    \centering
    \includegraphics[width=0.4\textwidth]{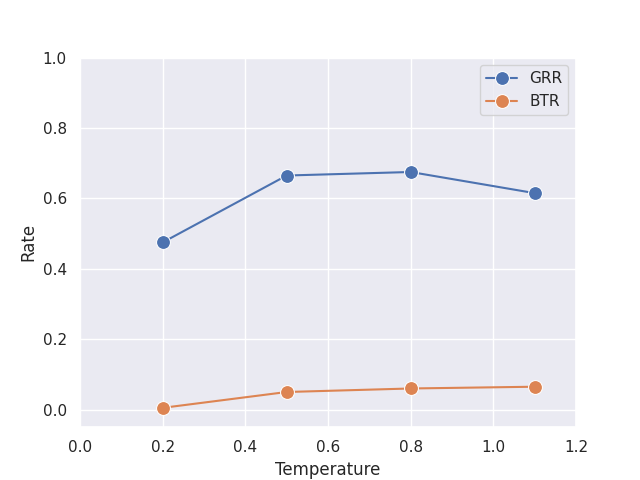}
    \caption{Plot comparing GRR of generator+verifier@25 with respect to varying values of temperature}
    \label{fig:GRR_temp}
\end{figure}

The sampling temperature $\tau$ of the LM controls the diversity in the output by scaling the un-normalized log probabilities of each token by a factor of $\tau$. A temperature of 0 would lead to greedy sampling and a temperature of infinity would lead to a uniform distribution over next tokens. To investigate the role of temperature in the performance of the architecture, we do a sweep over $\tau \in \{0.2, 0.5, 0.8, 1.1\}$. The results can be seen in Figure \ref{fig:GRR_temp}. 

The inverted U-shaped curve can be explained by the exploration-exploitation tradeoff where there is an optima reached at a particular temperature. At a lower temperature$(\tau=0.2)$, the generator is very conservative and doesn't have a lot of diversity in the generated plans. Suprisingly, it also makes a much lower number of bad transitions. On the other hand, at a higher temperature$(\tau=1.1)$, the generator is very exploratory and starts making more bad transitions, which is why the goal reaching rate starts decreasing. This is also suggestive of the fact that $\tau$ can be tuned on a validation set to find the optimal value. 

\section{Related Work}
Unlike methods such as SayCan \cite{saycan}, which use LLMs for heuristic guidance, and depend instead on external simulators for correctness, our verifier is learned from the same trajectory data used for finetuning the LLM in the first place. 
The idea of learning to act from an offline dataset of trajectories ties in to the field of offline RL. The ability to plan from the ``logs'' of a system provides efficiency by reducing the number of interactions required with the environment to learn the dynamics/policy in the environment. Even for PDDL planning, the efforts to leverage previous trajectories to provide planning capabilities pre-dates the rise of transformers. For example, \cite{hankz} use word vector embeddings and RNNs to support plan recognition. Recent works such as Decision Transformer\cite{dt} also employ large Transformer models to learn policies purely from trajectories. 
The work by \cite{plansformer} attempts to learn plans directly from action sequences generated by a planner. They too represent the states and actions in the form of text which acts as input for an LLM. 
Neither of these works however leverage  independently trained verifiers like we do. 

\section{Conclusions and Future Work}
From our study, we discover that the usage of a verifier is critical for using Large Language Models for the purposes of planning, since it can prune plans which have illegal actions. It is able to successfully unlock scaling with respect to an increasing number of attempts at generating a plan. An interesting question might be how to employ the feedback of verifiers into the generator during test-time. Also, how does the technique of verification scale to the level of pre-trained models as large as GPT-3 or GPT-4. Is it possible that a verifier could be built merely by prompt engineering via in-context learning? All these questions seem worth investigating, which we leave for future work. The appendix contains additional details on the training and test data, the hyper-parameter settings for the training of the generator and verifier, as well as an example of a 40 transition plan generated by the generator+verifier@25.

\clearpage
\bibliography{aaai23}

\section{Appendix}
\subsection{Training data}\label{subsec:Training_data}
The training set consists of 10000 initial states with 3 to 8 blocks. The generators used were taken from \cite{generators}. Using these, we generate trajectories of length at most 20, by sampling a random action from the valid set of actions in a state. Whenever a state is visited which has already been visited, we stop the trajectory. Then, we collect all the transitions in that trajectory with the goal being the end state. This allowed us to generate almost 61,000 trajectories. We create a similar validation dataset of 6000 trajectories. To convert the PDDL state and action to its textual representation, we convert all the true propositions to text and append them in lexicographical order.

\subsection{Test instances}
With a process similar to the training data generation, we generate 200 trajectories on different initial states of length atmost 30. The end state is sampled from the second half of the generated trajectory, so that the end state requires a non-trivial number of actions to reach. 

\subsection{Generator}
The generator is a finetuned GPT-2 model trained on 2 Nvidia V100 GPUs. The hyperparameters for training the generator are given in Table \ref{tab:gen_hparams}

\begin{table}[ht]
\centering
\begin{tabular}{cc}
\textbf{Parameter} & \textbf{Value} \\ \hline
Batch size         & 16             \\ \hline
Training epochs    & 20             \\ \hline
Learning rate      & 5e-6           \\ \hline
Warmup steps       & 50            
\end{tabular}
\caption{Training hyperparameters for the generator}
\label{tab:gen_hparams}
\end{table}

\subsection{Verifier}
The verifier is finetuned on the generator which itself is a GPT-2. It was trained on 1 Nvidia V100 GPU. The hyperparameters for training the verifier are given in Table \ref{tab:ver_hparams}

\begin{table}[ht]
\centering
\begin{tabular}{cc}
\textbf{Parameter} & \textbf{Value} \\ \hline
Batch size         & 8             \\ \hline
Training epochs    & 1             \\ \hline
Learning rate      & 5e-6           \\ \hline
Warmup steps       & 20            
\end{tabular}
\caption{Training hyperparameters for the verifier}
\label{tab:ver_hparams}
\end{table}

\subsection{An Example Plan}
The following is an example plan generated by generator+verifier@25. Note that the plan will not be (near-)optimal, because the training data doesn't use an (near-)optimal planner for trajectories. If (near-)optimal trajectories are used for training, the quality of plans will also be improved. \\
\textbf{Transition} 1:\\
\texttt{
GOAL:
(arm-empty)
(clear b1)
(clear b2)
(on b1 b3)
(on b3 b4)
(on-table b2)
(on-table b4)
\\\\
STATE:
(arm-empty)
(clear b1)
(on b1 b4)
(on b3 b2)
(on b4 b3)
(on-table b2)
\\\\
ACTION:
(unstack b1 b4)
\\\\
NEXT STATE:
(clear b4)
(holding b1)
(on b3 b2)
(on b4 b3)
(on-table b2)
}\\\\
\textbf{Transition} 2:\\
\texttt{
GOAL:
(arm-empty)
(clear b1)
(clear b2)
(on b1 b3)
(on b3 b4)
(on-table b2)
(on-table b4)
\\\\
STATE:
(clear b4)
(holding b1)
(on b3 b2)
(on b4 b3)
(on-table b2)
\\\\
ACTION:
(putdown b1)
\\\\
NEXT STATE:
(arm-empty)
(clear b1)
(clear b4)
(on b3 b2)
(on b4 b3)
(on-table b1)
(on-table b2)
}\\\\
\textbf{Transition} 3:\\
\texttt{
GOAL:
(arm-empty)
(clear b1)
(clear b2)
(on b1 b3)
(on b3 b4)
(on-table b2)
(on-table b4)
\\\\
STATE:
(arm-empty)
(clear b1)
(clear b4)
(on b3 b2)
(on b4 b3)
(on-table b1)
(on-table b2)
\\\\
ACTION:
(pickup b1)
\\\\
NEXT STATE:
(clear b4)
(holding b1)
(on b3 b2)
(on b4 b3)
(on-table b2)
}\\\\
\textbf{Transition} 4:\\
\texttt{
GOAL:
(arm-empty)
(clear b1)
(clear b2)
(on b1 b3)
(on b3 b4)
(on-table b2)
(on-table b4)
\\\\
STATE:
(clear b4)
(holding b1)
(on b3 b2)
(on b4 b3)
(on-table b2)
\\\\
ACTION:
(putdown b1)
\\\\
NEXT STATE:
(arm-empty)
(clear b1)
(clear b4)
(on b3 b2)
(on b4 b3)
(on-table b1)
(on-table b2)
}\\\\
\textbf{Transition} 5:\\
\texttt{
GOAL:
(arm-empty)
(clear b1)
(clear b2)
(on b1 b3)
(on b3 b4)
(on-table b2)
(on-table b4)
\\\\
STATE:
(arm-empty)
(clear b1)
(clear b4)
(on b3 b2)
(on b4 b3)
(on-table b1)
(on-table b2)
\\\\
ACTION:
(unstack b4 b3)
\\\\
NEXT STATE:
(clear b1)
(clear b3)
(holding b4)
(on b3 b2)
(on-table b1)
(on-table b2)
}\\\\
\textbf{Transition} 6:\\
\texttt{
GOAL:
(arm-empty)
(clear b1)
(clear b2)
(on b1 b3)
(on b3 b4)
(on-table b2)
(on-table b4)
\\\\
STATE:
(clear b1)
(clear b3)
(holding b4)
(on b3 b2)
(on-table b1)
(on-table b2)
\\\\
ACTION:
(stack b4 b1)
\\\\
NEXT STATE:
(arm-empty)
(clear b3)
(clear b4)
(on b3 b2)
(on b4 b1)
(on-table b1)
(on-table b2)
}\\\\
\textbf{Transition} 7:\\
\texttt{
GOAL:
(arm-empty)
(clear b1)
(clear b2)
(on b1 b3)
(on b3 b4)
(on-table b2)
(on-table b4)
\\\\
STATE:
(arm-empty)
(clear b3)
(clear b4)
(on b3 b2)
(on b4 b1)
(on-table b1)
(on-table b2)
\\\\
ACTION:
(unstack b3 b2)
\\\\
NEXT STATE:
(clear b2)
(clear b4)
(holding b3)
(on b4 b1)
(on-table b1)
(on-table b2)
}\\\\
\textbf{Transition} 8:\\
\texttt{
GOAL:
(arm-empty)
(clear b1)
(clear b2)
(on b1 b3)
(on b3 b4)
(on-table b2)
(on-table b4)
\\\\
STATE:
(clear b2)
(clear b4)
(holding b3)
(on b4 b1)
(on-table b1)
(on-table b2)
\\\\
ACTION:
(putdown b3)
\\\\
NEXT STATE:
(arm-empty)
(clear b2)
(clear b3)
(clear b4)
(on b4 b1)
(on-table b1)
(on-table b2)
(on-table b3)
}\\\\
\textbf{Transition} 9:\\
\texttt{
GOAL:
(arm-empty)
(clear b1)
(clear b2)
(on b1 b3)
(on b3 b4)
(on-table b2)
(on-table b4)
\\\\
STATE:
(arm-empty)
(clear b2)
(clear b3)
(clear b4)
(on b4 b1)
(on-table b1)
(on-table b2)
(on-table b3)
\\\\
ACTION:
(unstack b4 b1)
\\\\
NEXT STATE:
(clear b1)
(clear b2)
(clear b3)
(holding b4)
(on-table b1)
(on-table b2)
(on-table b3)
}\\\\
\textbf{Transition} 10:\\
\texttt{
GOAL:
(arm-empty)
(clear b1)
(clear b2)
(on b1 b3)
(on b3 b4)
(on-table b2)
(on-table b4)
\\\\
STATE:
(clear b1)
(clear b2)
(clear b3)
(holding b4)
(on-table b1)
(on-table b2)
(on-table b3)
\\\\
ACTION:
(putdown b4)
\\\\
NEXT STATE:
(arm-empty)
(clear b1)
(clear b2)
(clear b3)
(clear b4)
(on-table b1)
(on-table b2)
(on-table b3)
(on-table b4)
}\\\\
\textbf{Transition} 11:\\
\texttt{
GOAL:
(arm-empty)
(clear b1)
(clear b2)
(on b1 b3)
(on b3 b4)
(on-table b2)
(on-table b4)
\\\\
STATE:
(arm-empty)
(clear b1)
(clear b2)
(clear b3)
(clear b4)
(on-table b1)
(on-table b2)
(on-table b3)
(on-table b4)
\\\\
ACTION:
(pickup b2)
\\\\
NEXT STATE:
(clear b1)
(clear b3)
(clear b4)
(holding b2)
(on-table b1)
(on-table b3)
(on-table b4)
}\\\\
\textbf{Transition} 12:\\
\texttt{
GOAL:
(arm-empty)
(clear b1)
(clear b2)
(on b1 b3)
(on b3 b4)
(on-table b2)
(on-table b4)
\\\\
STATE:
(clear b1)
(clear b3)
(clear b4)
(holding b2)
(on-table b1)
(on-table b3)
(on-table b4)
\\\\
ACTION:
(stack b2 b1)
\\\\
NEXT STATE:
(arm-empty)
(clear b2)
(clear b3)
(clear b4)
(on b2 b1)
(on-table b1)
(on-table b3)
(on-table b4)
}\\\\
\textbf{Transition} 13:\\
\texttt{
GOAL:
(arm-empty)
(clear b1)
(clear b2)
(on b1 b3)
(on b3 b4)
(on-table b2)
(on-table b4)
\\\\
STATE:
(arm-empty)
(clear b2)
(clear b3)
(clear b4)
(on b2 b1)
(on-table b1)
(on-table b3)
(on-table b4)
\\\\
ACTION:
(unstack b2 b1)
\\\\
NEXT STATE:
(clear b1)
(clear b3)
(clear b4)
(holding b2)
(on-table b1)
(on-table b3)
(on-table b4)
}\\\\
\textbf{Transition} 14:\\
\texttt{
GOAL:
(arm-empty)
(clear b1)
(clear b2)
(on b1 b3)
(on b3 b4)
(on-table b2)
(on-table b4)
\\\\
STATE:
(clear b1)
(clear b3)
(clear b4)
(holding b2)
(on-table b1)
(on-table b3)
(on-table b4)
\\\\
ACTION:
(stack b2 b3)
\\\\
NEXT STATE:
(arm-empty)
(clear b1)
(clear b2)
(clear b4)
(on b2 b3)
(on-table b1)
(on-table b3)
(on-table b4)
}\\\\
\textbf{Transition} 15:\\
\texttt{
GOAL:
(arm-empty)
(clear b1)
(clear b2)
(on b1 b3)
(on b3 b4)
(on-table b2)
(on-table b4)
\\\\
STATE:
(arm-empty)
(clear b1)
(clear b2)
(clear b4)
(on b2 b3)
(on-table b1)
(on-table b3)
(on-table b4)
\\\\
ACTION:
(unstack b2 b3)
\\\\
NEXT STATE:
(clear b1)
(clear b3)
(clear b4)
(holding b2)
(on-table b1)
(on-table b3)
(on-table b4)
}\\\\
\textbf{Transition} 16:\\
\texttt{
GOAL:
(arm-empty)
(clear b1)
(clear b2)
(on b1 b3)
(on b3 b4)
(on-table b2)
(on-table b4)
\\\\
STATE:
(clear b1)
(clear b3)
(clear b4)
(holding b2)
(on-table b1)
(on-table b3)
(on-table b4)
\\\\
ACTION:
(stack b2 b4)
\\\\
NEXT STATE:
(arm-empty)
(clear b1)
(clear b2)
(clear b3)
(on b2 b4)
(on-table b1)
(on-table b3)
(on-table b4)
}\\\\
\textbf{Transition} 17:\\
\texttt{
GOAL:
(arm-empty)
(clear b1)
(clear b2)
(on b1 b3)
(on b3 b4)
(on-table b2)
(on-table b4)
\\\\
STATE:
(arm-empty)
(clear b1)
(clear b2)
(clear b3)
(on b2 b4)
(on-table b1)
(on-table b3)
(on-table b4)
\\\\
ACTION:
(pickup b3)
\\\\
NEXT STATE:
(clear b1)
(clear b2)
(holding b3)
(on b2 b4)
(on-table b1)
(on-table b4)
}\\\\
\textbf{Transition} 18:\\
\texttt{
GOAL:
(arm-empty)
(clear b1)
(clear b2)
(on b1 b3)
(on b3 b4)
(on-table b2)
(on-table b4)
\\\\
STATE:
(clear b1)
(clear b2)
(holding b3)
(on b2 b4)
(on-table b1)
(on-table b4)
\\\\
ACTION:
(stack b3 b2)
\\\\
NEXT STATE:
(arm-empty)
(clear b1)
(clear b3)
(on b2 b4)
(on b3 b2)
(on-table b1)
(on-table b4)
}\\\\
\textbf{Transition} 19:\\
\texttt{
GOAL:
(arm-empty)
(clear b1)
(clear b2)
(on b1 b3)
(on b3 b4)
(on-table b2)
(on-table b4)
\\\\
STATE:
(arm-empty)
(clear b1)
(clear b3)
(on b2 b4)
(on b3 b2)
(on-table b1)
(on-table b4)
\\\\
ACTION:
(unstack b3 b2)
\\\\
NEXT STATE:
(clear b1)
(clear b2)
(holding b3)
(on b2 b4)
(on-table b1)
(on-table b4)
}\\\\
\textbf{Transition} 20:\\
\texttt{
GOAL:
(arm-empty)
(clear b1)
(clear b2)
(on b1 b3)
(on b3 b4)
(on-table b2)
(on-table b4)
\\\\
STATE:
(clear b1)
(clear b2)
(holding b3)
(on b2 b4)
(on-table b1)
(on-table b4)
\\\\
ACTION:
(putdown b3)
\\\\
NEXT STATE:
(arm-empty)
(clear b1)
(clear b2)
(clear b3)
(on b2 b4)
(on-table b1)
(on-table b3)
(on-table b4)
}\\\\
\textbf{Transition} 21:\\
\texttt{
GOAL:
(arm-empty)
(clear b1)
(clear b2)
(on b1 b3)
(on b3 b4)
(on-table b2)
(on-table b4)
\\\\
STATE:
(arm-empty)
(clear b1)
(clear b2)
(clear b3)
(on b2 b4)
(on-table b1)
(on-table b3)
(on-table b4)
\\\\
ACTION:
(pickup b3)
\\\\
NEXT STATE:
(clear b1)
(clear b2)
(holding b3)
(on b2 b4)
(on-table b1)
(on-table b4)
}\\\\
\textbf{Transition} 22:\\
\texttt{
GOAL:
(arm-empty)
(clear b1)
(clear b2)
(on b1 b3)
(on b3 b4)
(on-table b2)
(on-table b4)
\\\\
STATE:
(clear b1)
(clear b2)
(holding b3)
(on b2 b4)
(on-table b1)
(on-table b4)
\\\\
ACTION:
(putdown b3)
\\\\
NEXT STATE:
(arm-empty)
(clear b1)
(clear b2)
(clear b3)
(on b2 b4)
(on-table b1)
(on-table b3)
(on-table b4)
}\\\\
\textbf{Transition} 23:\\
\texttt{
GOAL:
(arm-empty)
(clear b1)
(clear b2)
(on b1 b3)
(on b3 b4)
(on-table b2)
(on-table b4)
\\\\
STATE:
(arm-empty)
(clear b1)
(clear b2)
(clear b3)
(on b2 b4)
(on-table b1)
(on-table b3)
(on-table b4)
\\\\
ACTION:
(pickup b1)
\\\\
NEXT STATE:
(clear b2)
(clear b3)
(holding b1)
(on b2 b4)
(on-table b3)
(on-table b4)
}\\\\
\textbf{Transition} 24:\\
\texttt{
GOAL:
(arm-empty)
(clear b1)
(clear b2)
(on b1 b3)
(on b3 b4)
(on-table b2)
(on-table b4)
\\\\
STATE:
(clear b2)
(clear b3)
(holding b1)
(on b2 b4)
(on-table b3)
(on-table b4)
\\\\
ACTION:
(putdown b1)
\\\\
NEXT STATE:
(arm-empty)
(clear b1)
(clear b2)
(clear b3)
(on b2 b4)
(on-table b1)
(on-table b3)
(on-table b4)
}\\\\
\textbf{Transition} 25:\\
\texttt{
GOAL:
(arm-empty)
(clear b1)
(clear b2)
(on b1 b3)
(on b3 b4)
(on-table b2)
(on-table b4)
\\\\
STATE:
(arm-empty)
(clear b1)
(clear b2)
(clear b3)
(on b2 b4)
(on-table b1)
(on-table b3)
(on-table b4)
\\\\
ACTION:
(pickup b3)
\\\\
NEXT STATE:
(clear b1)
(clear b2)
(holding b3)
(on b2 b4)
(on-table b1)
(on-table b4)
}\\\\
\textbf{Transition} 26:\\
\texttt{
GOAL:
(arm-empty)
(clear b1)
(clear b2)
(on b1 b3)
(on b3 b4)
(on-table b2)
(on-table b4)
\\\\
STATE:
(clear b1)
(clear b2)
(holding b3)
(on b2 b4)
(on-table b1)
(on-table b4)
\\\\
ACTION:
(putdown b3)
\\\\
NEXT STATE:
(arm-empty)
(clear b1)
(clear b2)
(clear b3)
(on b2 b4)
(on-table b1)
(on-table b3)
(on-table b4)
}\\\\
\textbf{Transition} 27:\\
\texttt{
GOAL:
(arm-empty)
(clear b1)
(clear b2)
(on b1 b3)
(on b3 b4)
(on-table b2)
(on-table b4)
\\\\
STATE:
(arm-empty)
(clear b1)
(clear b2)
(clear b3)
(on b2 b4)
(on-table b1)
(on-table b3)
(on-table b4)
\\\\
ACTION:
(unstack b2 b4)
\\\\
NEXT STATE:
(clear b1)
(clear b3)
(clear b4)
(holding b2)
(on-table b1)
(on-table b3)
(on-table b4)
}\\\\
\textbf{Transition} 28:\\
\texttt{
GOAL:
(arm-empty)
(clear b1)
(clear b2)
(on b1 b3)
(on b3 b4)
(on-table b2)
(on-table b4)
\\\\
STATE:
(clear b1)
(clear b3)
(clear b4)
(holding b2)
(on-table b1)
(on-table b3)
(on-table b4)
\\\\
ACTION:
(stack b2 b1)
\\\\
NEXT STATE:
(arm-empty)
(clear b2)
(clear b3)
(clear b4)
(on b2 b1)
(on-table b1)
(on-table b3)
(on-table b4)
}\\\\
\textbf{Transition} 29:\\
\texttt{
GOAL:
(arm-empty)
(clear b1)
(clear b2)
(on b1 b3)
(on b3 b4)
(on-table b2)
(on-table b4)
\\\\
STATE:
(arm-empty)
(clear b2)
(clear b3)
(clear b4)
(on b2 b1)
(on-table b1)
(on-table b3)
(on-table b4)
\\\\
ACTION:
(unstack b2 b1)
\\\\
NEXT STATE:
(clear b1)
(clear b3)
(clear b4)
(holding b2)
(on-table b1)
(on-table b3)
(on-table b4)
}\\\\
\textbf{Transition} 30:\\
\texttt{
GOAL:
(arm-empty)
(clear b1)
(clear b2)
(on b1 b3)
(on b3 b4)
(on-table b2)
(on-table b4)
\\\\
STATE:
(clear b1)
(clear b3)
(clear b4)
(holding b2)
(on-table b1)
(on-table b3)
(on-table b4)
\\\\
ACTION:
(stack b2 b3)
\\\\
NEXT STATE:
(arm-empty)
(clear b1)
(clear b2)
(clear b4)
(on b2 b3)
(on-table b1)
(on-table b3)
(on-table b4)
}\\\\
\textbf{Transition} 31:\\
\texttt{
GOAL:
(arm-empty)
(clear b1)
(clear b2)
(on b1 b3)
(on b3 b4)
(on-table b2)
(on-table b4)
\\\\
STATE:
(arm-empty)
(clear b1)
(clear b2)
(clear b4)
(on b2 b3)
(on-table b1)
(on-table b3)
(on-table b4)
\\\\
ACTION:
(unstack b2 b3)
\\\\
NEXT STATE:
(clear b1)
(clear b3)
(clear b4)
(holding b2)
(on-table b1)
(on-table b3)
(on-table b4)
}\\\\
\textbf{Transition} 32:\\
\texttt{
GOAL:
(arm-empty)
(clear b1)
(clear b2)
(on b1 b3)
(on b3 b4)
(on-table b2)
(on-table b4)
\\\\
STATE:
(clear b1)
(clear b3)
(clear b4)
(holding b2)
(on-table b1)
(on-table b3)
(on-table b4)
\\\\
ACTION:
(stack b2 b1)
\\\\
NEXT STATE:
(arm-empty)
(clear b2)
(clear b3)
(clear b4)
(on b2 b1)
(on-table b1)
(on-table b3)
(on-table b4)
}\\\\
\textbf{Transition} 33:\\
\texttt{
GOAL:
(arm-empty)
(clear b1)
(clear b2)
(on b1 b3)
(on b3 b4)
(on-table b2)
(on-table b4)
\\\\
STATE:
(arm-empty)
(clear b2)
(clear b3)
(clear b4)
(on b2 b1)
(on-table b1)
(on-table b3)
(on-table b4)
\\\\
ACTION:
(pickup b3)
\\\\
NEXT STATE:
(clear b2)
(clear b4)
(holding b3)
(on b2 b1)
(on-table b1)
(on-table b4)
}\\\\
\textbf{Transition} 34:\\
\texttt{
GOAL:
(arm-empty)
(clear b1)
(clear b2)
(on b1 b3)
(on b3 b4)
(on-table b2)
(on-table b4)
\\\\
STATE:
(clear b2)
(clear b4)
(holding b3)
(on b2 b1)
(on-table b1)
(on-table b4)
\\\\
ACTION:
(stack b3 b4)
\\\\
NEXT STATE:
(arm-empty)
(clear b2)
(clear b3)
(on b2 b1)
(on b3 b4)
(on-table b1)
(on-table b4)
}\\\\
\textbf{Transition} 35:\\
\texttt{
GOAL:
(arm-empty)
(clear b1)
(clear b2)
(on b1 b3)
(on b3 b4)
(on-table b2)
(on-table b4)
\\\\
STATE:
(arm-empty)
(clear b2)
(clear b3)
(on b2 b1)
(on b3 b4)
(on-table b1)
(on-table b4)
\\\\
ACTION:
(unstack b2 b1)
\\\\
NEXT STATE:
(clear b1)
(clear b3)
(holding b2)
(on b3 b4)
(on-table b1)
(on-table b4)
}\\\\
\textbf{Transition} 36:\\
\texttt{
GOAL:
(arm-empty)
(clear b1)
(clear b2)
(on b1 b3)
(on b3 b4)
(on-table b2)
(on-table b4)
\\\\
STATE:
(clear b1)
(clear b3)
(holding b2)
(on b3 b4)
(on-table b1)
(on-table b4)
\\\\
ACTION:
(stack b2 b3)
\\\\
NEXT STATE:
(arm-empty)
(clear b1)
(clear b2)
(on b2 b3)
(on b3 b4)
(on-table b1)
(on-table b4)
}\\\\
\textbf{Transition} 37:\\
\texttt{
GOAL:
(arm-empty)
(clear b1)
(clear b2)
(on b1 b3)
(on b3 b4)
(on-table b2)
(on-table b4)
\\\\
STATE:
(arm-empty)
(clear b1)
(clear b2)
(on b2 b3)
(on b3 b4)
(on-table b1)
(on-table b4)
\\\\
ACTION:
(unstack b2 b3)
\\\\
NEXT STATE:
(clear b1)
(clear b3)
(holding b2)
(on b3 b4)
(on-table b1)
(on-table b4)
}\\\\
\textbf{Transition} 38:\\
\texttt{
GOAL:
(arm-empty)
(clear b1)
(clear b2)
(on b1 b3)
(on b3 b4)
(on-table b2)
(on-table b4)
\\\\
STATE:
(clear b1)
(clear b3)
(holding b2)
(on b3 b4)
(on-table b1)
(on-table b4)
\\\\
ACTION:
(putdown b2)
\\\\
NEXT STATE:
(arm-empty)
(clear b1)
(clear b2)
(clear b3)
(on b3 b4)
(on-table b1)
(on-table b2)
(on-table b4)
}\\\\
\textbf{Transition} 39:\\
\texttt{
GOAL:
(arm-empty)
(clear b1)
(clear b2)
(on b1 b3)
(on b3 b4)
(on-table b2)
(on-table b4)
\\\\
STATE:
(arm-empty)
(clear b1)
(clear b2)
(clear b3)
(on b3 b4)
(on-table b1)
(on-table b2)
(on-table b4)
\\\\
ACTION:
(pickup b1)
\\\\
NEXT STATE:
(clear b2)
(clear b3)
(holding b1)
(on b3 b4)
(on-table b2)
(on-table b4)
}\\\\
\textbf{Transition} 40:\\
\texttt{
GOAL:
(arm-empty)
(clear b1)
(clear b2)
(on b1 b3)
(on b3 b4)
(on-table b2)
(on-table b4)
\\\\
STATE:
(clear b2)
(clear b3)
(holding b1)
(on b3 b4)
(on-table b2)
(on-table b4)
\\\\
ACTION:
(stack b1 b3)
\\\\
NEXT STATE:
(arm-empty)
(clear b1)
(clear b2)
(on b1 b3)
(on b3 b4)
(on-table b2)
(on-table b4)
}\\\\

\end{document}